\documentclass[10pt,twocolumn,letterpaper]{article}

\usepackage[pagenumbers]{iccv} % To force page numbers, e.g. for an arXiv version

\usepackage{graphicx}
\usepackage{amsmath}
\usepackage{amssymb}
\usepackage{booktabs}
\usepackage{blindtext}
\usepackage{array}
\usepackage{graphicx}
\usepackage{multirow}
\usepackage{booktabs}
\usepackage{wrapfig}
\usepackage{caption}
\usepackage{diagbox}
\usepackage{adjustbox}
\usepackage{float}
\usepackage{xcolor}
\usepackage{colortbl}
\usepackage{comment}
\usepackage{multicol}
\usepackage{booktabs}  
\usepackage{threeparttable}
\usepackage{amsfonts}       % blackboard math symbols    
\usepackage{nicefrac}       % compact symbols for 1/2, etc.
\usepackage{microtype}      % microtypography
\usepackage{xcolor}         % colors
\usepackage{bm}     
\usepackage{xfrac} 
\usepackage{array}
\usepackage{makecell}
\usepackage{wrapfig}
\usepackage{enumitem}

\newcommand{\PreserveBackslash}[1]{\let\temp=\\#1\let\\=\temp}
\newcolumntype{C}[1]{>{\PreserveBackslash\centering}p{#1}}
\newcolumntype{R}[1]{>{\PreserveBackslash\raggedleft}p{#1}}
\newcolumntype{L}[1]{>{\PreserveBackslash\raggedright}p{#1}}

\definecolor{iccvblue}{rgb}{0.21,0.49,0.74}
\usepackage[pagebackref,breaklinks,colorlinks,allcolors=iccvblue]{hyperref}
\usepackage[T1]{fontenc}

\def\confName{ICCV}
\def\confYear{2025}

\title{Task-Aware Image Signal Processor for Advanced Visual Perception}

\author{
\hspace{-0.5cm}
Kai Chen\textsuperscript{1}\quad Jin Xiao\textsuperscript{2}\quad
Leheng Zhang\textsuperscript{1}\quad Kexuan Shi\textsuperscript{1} \quad Shuhang Gu\textsuperscript{1}\footnotemark[1]\\%\Letter
\hspace{-0.5cm}
 \textsuperscript{1}{University of Electronic Science and Technology of China} \hspace{0pt}\quad
 \textsuperscript{2}{ByteDance Inc.} \hspace{0pt}\\
\hspace{-0.5cm}
{\tt \small \{kkchen318, shuhanggu\}@gmail.com}\\
\small \url{https://github.com/LabShuHangGU/TA-ISP}
}

\begin{document}
\maketitle

\renewcommand{\thefootnote}{\fnsymbol{footnote}} 
\footnotetext[1]{Corresponding author.}

\begin{abstract}
In recent years, there has been a growing trend in computer vision towards exploiting RAW sensor data, which preserves richer information compared to conventional low-bit RGB images. Early studies mainly focused on enhancing visual quality, while more recent efforts aim to leverage the abundant information in RAW data to improve the performance of visual perception tasks such as object detection and segmentation. However, existing approaches still face two key limitations: large-scale ISP networks impose heavy computational overhead, while methods based on tuning traditional ISP pipelines are restricted by limited representational capacity.
To address these issues, we propose \textbf{T}ask-\textbf{A}ware \textbf{I}mage \textbf{S}ignal \textbf{P}rocessing (\textbf{TA-ISP})
, a compact RAW-to-RGB framework that produces task-oriented representations for pretrained vision models. Instead of heavy dense convolutional pipelines, TA-ISP predicts a small set of lightweight, multi-scale modulation operators that act at global, regional, and pixel scales to reshape image statistics across different spatial extents. This factorized control significantly expands the range of spatially varying transforms that can be represented while keeping memory usage, computation, and latency tightly constrained. Evaluated on several RAW-domain detection and segmentation benchmarks under both daytime and nighttime conditions, TA-ISP consistently improves downstream accuracy while markedly reducing parameter count and inference time, making it well suited for deployment on resource-constrained devices.
\end{abstract}    
\section{Introduction}
\label{sec:intro}

The past decades have witnessed remarkable success in various computer vision applications.
While early works \cite{yolox2021, xie2021segformer, matterport_maskrcnn_2017} rely on RGB data as input, focusing on performance improvement through architectural designs and training strategies, a recent trend has emerged towards utilizing RAW data for visual tasks, as it preserves more information than low-bit RGB images.

Early works \cite{ignatov2020aim2020challengelearned,chen2018learningdark} on RAW data processing primarily focused on enhancing the visual quality of images. 
These efforts leveraged the powerful nonlinear fitting capabilities of neural networks, along with large datasets of RAW-RGB images, to produce high-quality imaging results from low-quality RAW inputs. 
Over the past few years, researchers have increasingly focused on how to fully utilize the rich information contained in RAW data to improve the performance of visual perception tasks \cite{Wu_2019,flexisp,yoshimura2023dynamicispdynamicallycontrolledimage,diap,yu2021reconfigispreconfigurablecameraimage} such as detection and segmentation. 
By either tuning the parameters of traditional image signal processor (ISP) for specific downstream tasks \cite{refactor,yu2021reconfigispreconfigurablecameraimage,wang2024adaptiveisplearningadaptiveimage,cui2024rawadapteradaptingpretrainedvisual} or jointly optimizing an ISP network with a downstream vision model \cite{yoshimura2023dynamicispdynamicallycontrolledimage,diap,steven:dirtypixels2021}
, models based on RAW data have shown promising results in various vision applications.

\begin{figure}[t]
    \centering
    \includegraphics[width=0.49 \textwidth]{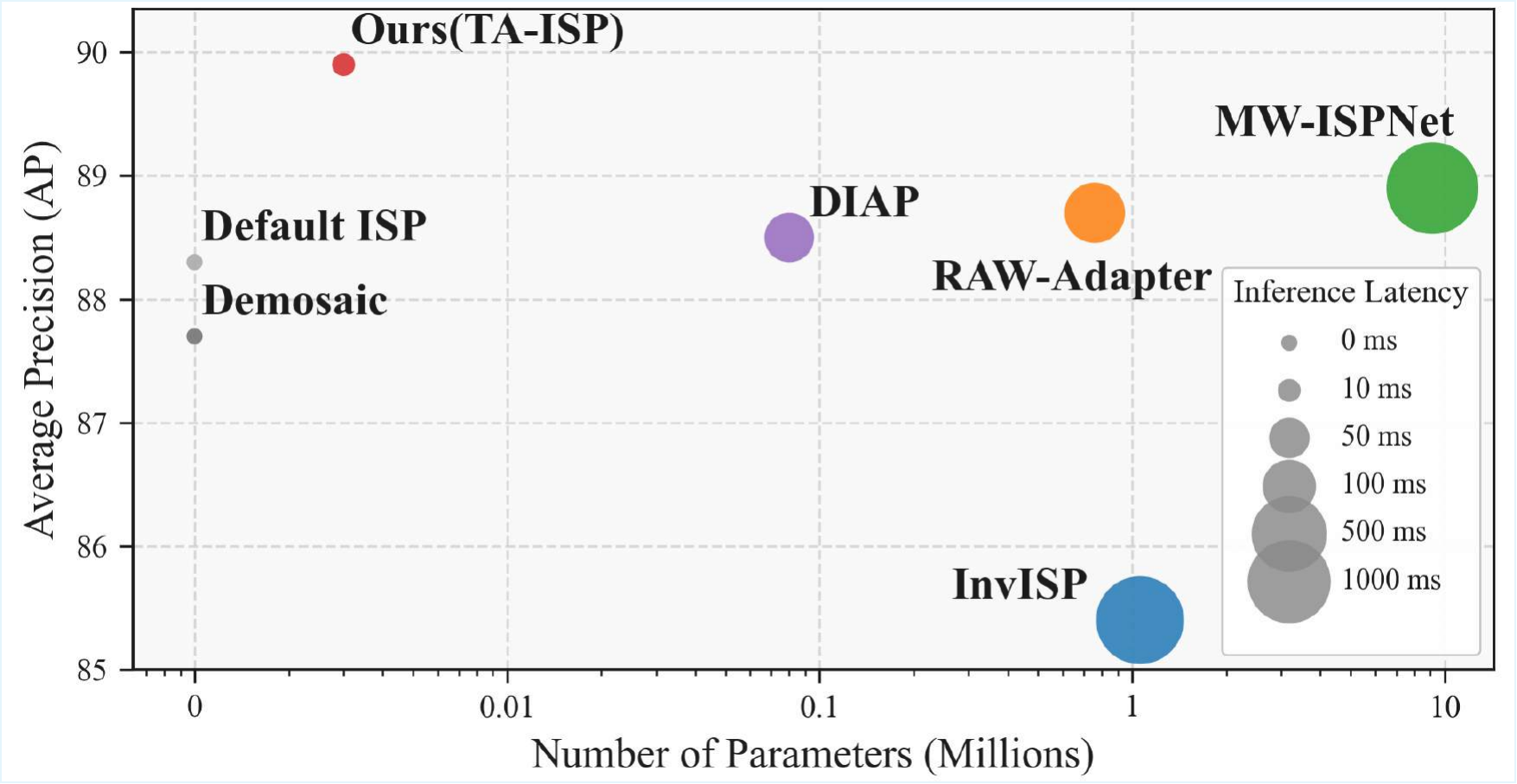}
    \captionsetup{width=0.475\textwidth}
    \caption{Performance comparison of different methods in terms of inference latency, parameter count, and average precision. Among all approaches, TA-ISP consistently achieves the best performance across all metrics, significantly surpassing existing methods.}
    \label{performance}
    \vspace{-1em}
\end{figure}

Despite their success on benchmark datasets, 
existing RAW-based methods still face two major challenges.
The first challenge lies in the computational and data transmission costs.
Current approaches often rely on large-scale neural networks with heavy computational footprints to learn RAW-to-RGB mappings \cite{ignatov2020aim2020challengelearned,steven:dirtypixels2021} or to inject additional intermediate information into downstream models \cite{cui2024rawadapteradaptingpretrainedvisual}, introducing substantial burdens that can degrade the performance of terminal devices. However, ISP is fundamentally required to be constrained in terms of area and power consumption, while still playing a critical role in promptly reducing data bit-width on terminal devices to lower transmission costs.
Secondly, many existing works \cite{cui2024rawadapteradaptingpretrainedvisual, refactor, yu2021reconfigispreconfigurablecameraimage, wang2024adaptiveisplearningadaptiveimage} adopt a lightweight strategy that tunes only a small set of parameters within a conventional ISP in order to better align the pipeline with downstream tasks. Although this approach is appealing due to its low computational and implementation cost, its expressive capacity remains fundamentally limited: adjusting only global or per-channel parameters alone cannot adequately capture the complex, spatially varying transformations required in practice. As a result, such parameter-tuning methods often fail to generalize to unseen scenes or tasks when the necessary transformations extend beyond the design space of the original ISP.

To address these limitations, we propose \textbf{TA-ISP}, a task-aware, lightweight RAW-to-RGB pipeline that explicitly learns to produce representations optimized for downstream vision models while remaining deployment-friendly. TA-ISP departs from both hand-tuned, low-capacity ISP adjustments and heavy end-to-end ISP networks by introducing a suite of multi-granularity pixel-modulation modules operating at global, regional, and pixel levels. Owing to this factorized design, most of the computation is spent on predicting compact operations at different scales rather than on dense convolutions, keeping the computational burden low. At the same time, the combination of global, regional, and pixel modulations enables a rich family of spatially-varying transforms. This hierarchical decomposition substantially enlarges the representational space compared with parameter-tuning ISPs, while avoiding the cost of monolithic deep networks, thereby offering both lightweight efficiency and strong adaptability to diverse scenes and downstream tasks. We validate TA-ISP on representative perception benchmarks and demonstrate that it consistently delivers superior accuracy while maintaining a compact parameter footprint and inference latency, as shown in Figure \ref{performance}.

Our contributions can be summarized as follows:

\begin{itemize}
    \item We present TA-ISP, a lightweight and deployment-friendly RAW-to-RGB framework that attains task-oriented performance gains while satisfying strict constraints on model size, inference latency, and data transmission.
    \item We design and integrate multi-granularity pixel-modulation modules (global-level, regional-level, and pixel-level) that jointly adapt image statistics at different spatial scales to produce compact and task-adaptive RAW-to-RGB representations.
    \item We conduct extensive experiments on RAW-domain detection and segmentation benchmarks under both daytime and nighttime conditions. Compared with various mainstream methods, our TA-ISP consistently and significantly achieves state-of-the-art (SOTA) performance.
\end{itemize}

\section{Related Works}
\label{sec:formatting}

% All text must be in a two-column format.
% The total allowable size of the text area is $6\frac78$ inches (17.46 cm) wide by $8\frac78$ inches (22.54 cm) high.
% Columns are to be $3\frac14$ inches (8.25 cm) wide, with a $\frac{5}{16}$ inch (0.8 cm) space between them.
% The main title (on the first page) should begin 1 inch (2.54 cm) from the top edge of the page.
% The second and following pages should begin 1 inch (2.54 cm) from the top edge.
% On all pages, the bottom margin should be $1\frac{1}{8}$ inches (2.86 cm) from the bottom edge of the page for $8.5 \times 11$-inch paper;
% for A4 paper, approximately $1\frac{5}{8}$ inches (4.13 cm) from the bottom edge of the
% page.

%-------------------------------------------------------------------------
\subsection{Image Signal Processor}

Image Signal Processing (ISP) aims to produce high-quality RGB images from the RAW images captured by sensors. The traditional ISP pipeline generally consists of a series of steps  \citep{KaraimerB16,nishimura2019automaticispimagequality,1407713,delbracio2021mobilecomputationalphotographytour}, including demosaicing, denoising, white balance, gamma correction, and color transformation. Some works attempt to replace the traditional ISP by alternative design \cite{Hasinoff,flexisp}. Heide \textit{et al.} \cite{flexisp} identified that the traditional "divide and conquer" approach of ISP can introduce cumulative errors and proposed to merge the ISP pipeline as a solution. Hasinoff \textit{et al.} \cite{Hasinoff} revised the traditional ISP steps to address the challenges of extremely low-light photography.

With the advancement of deep learning, many approaches have sought to replace part or all of the classical ISP with learned end-to-end models that map RAW-to-RGB \cite{xing2021invertibleimagesignalprocessing,chen2018learningdark,ignatov2020aim2020challengelearned,Conde_2022,replace,kim2024paramisplearnedforwardinverse,zhang2021learningrawtosrgbmappingsinaccurately,9151012,liang2019cameranettwostageframeworkeffective,Schwartz_2019}. For instance, wavelet-based UNet variants have been introduced to better preserve high-frequency details during reconstruction \cite{ignatov2020aim2020challengelearned}, while invertible architectures aim to explicitly learn both forward and inverse mappings between RAW and RGB domains \cite{xing2021invertibleimagesignalprocessing}. However, such deep network models are often large and computationally intensive, which limits their practicality in real-world deployment. Beyond RAW–RGB conversion, an emerging line of research instead focuses on directly leveraging RAW data to enhance downstream vision tasks.

\begin{figure*}[htbp]
    \centering
    \includegraphics[width=1.0\textwidth]{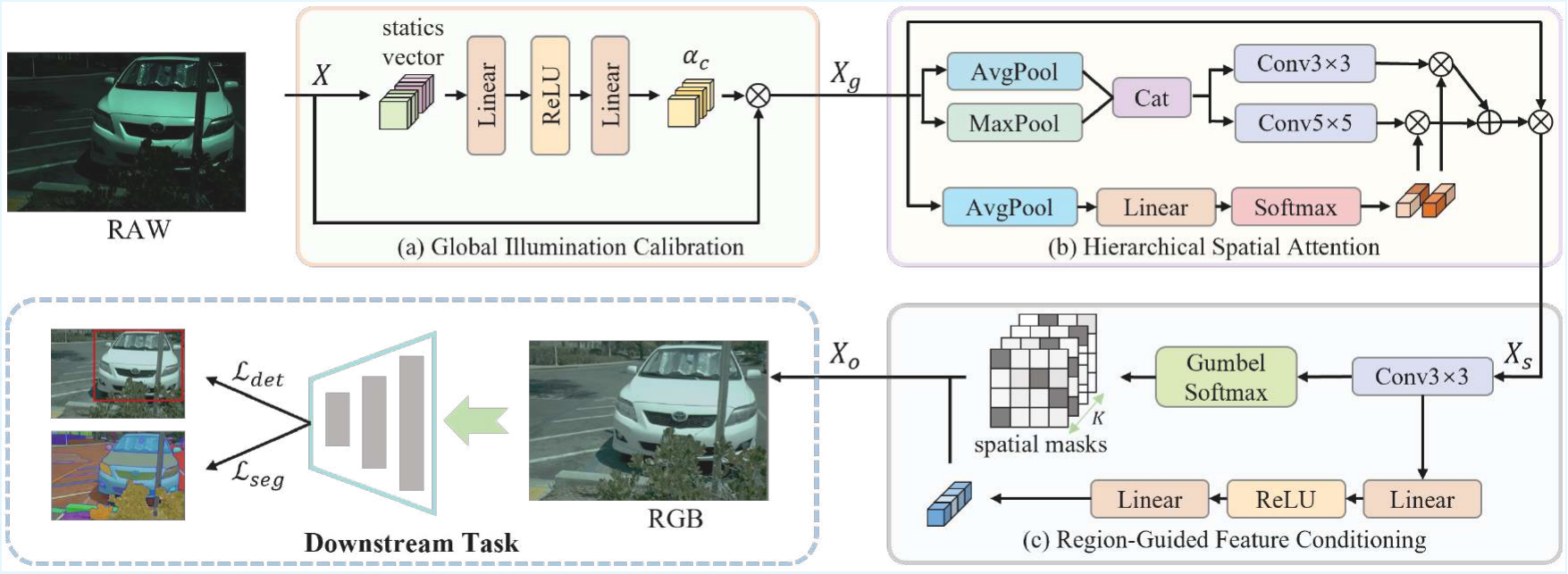} 
    \caption{The overall framework of TA-ISP. (a). Global Illumination Calibration. (b). Hierarchical Spatial Attention. (c). Region-Guided Feature Conditioning. These three modules progressively adjust the representations at the global, regional, and pixel levels, thereby facilitating more effective adaptation to downstream models. }
    \label{framework}
    % \vspace{-1em}
\end{figure*}

\subsection{Visual Perception on RAW data}

To fully exploit the advantages of RAW data, several approaches \cite{Wu_2019, cui2024rawadapteradaptingpretrainedvisual,yu2021reconfigispreconfigurablecameraimage, yoshimura2023dynamicispdynamicallycontrolledimage,steven:dirtypixels2021,attention, guo2024learningdegradationindependentrepresentationscamera,wang2024adaptiveisplearningadaptiveimage} have explored using RAW images for downstream visual perception tasks. VisionISP \cite{Wu_2019} was among the first to highlight the discrepancy between human and machine vision and introduced trainable modules to enhance task performance. Following this, many works have focused on designing differentiable network architectures for end-to-end joint optimization \cite{steven:dirtypixels2021,hardware,yoshimura2023dynamicispdynamicallycontrolledimage}.
However, training large-scale ISP networks inevitably imposes substantial computational burdens on resource-constrained devices.
An alternative direction focuses on lightweight adaptation of classical ISP pipelines by tuning a small set of parameters or inserting compact adapter modules \cite{yu2021reconfigispreconfigurablecameraimage,refactor,cui2024rawadapteradaptingpretrainedvisual,wang2024adaptiveisplearningadaptiveimage,diap}. For instance, DIAP \cite{diap} underscores the importance of high dynamic range for downstream tasks and introduces lightweight adjustment modules to better align the ISP with task requirements, while RAW-Adapter \cite{cui2024rawadapteradaptingpretrainedvisual} fuses features from a conventional ISP into a downstream model and updates a modest number of parameters. Such approaches are appealing due to their low computational overhead and ease of deployment, yet their representational capacity remains fundamentally constrained, making it difficult to fully accommodate the diverse and spatially complex requirements of downstream tasks.
In this work, we propose a task-aware ISP (TA-ISP) that adapts to downstream models through a set of multi-granularity pixel-adjustment modules. By operating at different spatial scales, these modules enable the ISP to learn task-adaptive representations in conjunction with downstream networks, thereby achieving superior performance compared to existing methods.

\section{Proposed Method}
In this section, we begin by discussing the motivation that underlies our approach. We then provide a concrete illustration of our network architecture. The overall framework is shown in Figure \ref{framework}.

\subsection{Motivation}
%加强表征
Visual perception aims to automatically extract and analyze useful information from visual data.
Most existing visual perception studies \cite{yolox2021,xie2021segformer,matterport_maskrcnn_2017} simply take off-the-shelf processed RGB images as input, focusing primarily on designing vision algorithms for RGB data.
However, the separately designed ISP function cannot be optimized for the specific requirements of different downstream tasks, often resulting in suboptimal representation of visual information and limiting the performance of subsequent models.

Recently, to fully exploit the high-bit and information-rich RAW data, joint optimization methods \cite{steven:dirtypixels2021,cui2024rawadapteradaptingpretrainedvisual,diap,yu2021reconfigispreconfigurablecameraimage} have been proposed to establish advanced RAW-to-RGB mapping pipelines tailored to downstream task requirements.
Despite achieving superior perceptual results on benchmark datasets, these methods either incur heavy computational costs \cite{steven:dirtypixels2021} or merely adjust a limited set of parameters in traditional ISPs \cite{cui2024rawadapteradaptingpretrainedvisual,yu2021reconfigispreconfigurablecameraimage}, limiting their practicality for resource-constrained terminal devices.

Motivated by \cite{diap}, which highlights the importance of high dynamic range for task-oriented imaging, we focus on adapting the distribution of high-bit RAW measurements to better serve downstream vision models. To this end, we propose task-aware ISP (TA-ISP), a lightweight RAW-to-RGB pipeline that produces task-specific representations directly from RAW inputs. TA-ISP introduces a set of multi-granularity pixel-modulation modules that adjust exposure, color, and spatial responses at global, regional, and pixel levels. This design enables the network to capture rich, spatially-varying transformations required by downstream models, while keeping computation centered on compact adjustment operations rather than heavy convolutions. As a result, TA-ISP delivers task-aware representations with both high adaptability and low computational cost, offering a lightweight yet effective RAW-to-RGB processing solution.
In the following subsections \ref{nano}, we will introduce details of our network architecture.

\subsection{TA-ISP}\label{nano}
\subsubsection{Overview}
%加强表征
In this section, we present the architecture of our task-aware ISP network, which is tailored to efficiently process RAW sensor data while maintaining essential information for downstream tasks.
An illustration of the network architecture of our task-aware ISP can be found in Figure \ref{framework}. Our network is jointly optimized with the downstream vision model in an end-to-end manner, enabling it to adaptively enhance RAW sensor inputs for task-specific requirements. 
Specifically, our network mainly comprises three components, i.e. the global luminance calibration module, the hierarchical spatial attention module and the region-guided feature conditioning module. 
The RAW data is first processed by the global luminance calibration module, which performs coarse adjustments on exposure and color to ensure balanced brightness across channels. The output is then sequentially passed through the hierarchical spatial attention module, capturing multi-scale spatial dependencies and enabling pixel-level attention to salient structures while suppressing irrelevant regions. Finally, the region-guided feature conditioning module applies spatial masks and channel-wise weights in a region-aware manner, highlighting content critical for downstream tasks. The output of the final module is the processed RGB image, which is then directly fed to the downstream vision model.
This design allows our ISP network to efficiently enhance task-specific representations in RAW data, providing a lightweight and adaptive solution for RAW-to-RGB processing.

\subsubsection{Global Luminance Calibration.}
In visual perception tasks with RAW sensor inputs, pixel values are often concentrated within a narrow low-intensity range, which limits the expressive capacity of learned representations and diminishes their usefulness for downstream tasks. Moreover, inherent sensor characteristics  lead to pronounced channel-wise exposure imbalance, resulting in biased color responses that can further hinder feature learning and reduce task performance.

To address the limited dynamic range and channel-wise imbalance inherent in RAW measurements, we introduce a lightweight global luminance calibration module. 
Given a packed RAW tensor $X \in \mathbb{R}^{C \times H \times W}$, We first compute the global statistics of the image, specifically the mean and variance for each channel:
\begin{equation}
    \mu_c = \frac{1}{HW} \sum_{i,j} X_{c,i,j}, \quad
\sigma_c^2 = \frac{1}{HW} \sum_{i,j} (X_{c,i,j} - \mu_c)^2.
\end{equation}
These statistics are concatenated and fed into fully-connected layers to estimate per-channel multiplicative gains $\alpha_c$:
\begin{equation}
   \alpha_c = \text{Softplus}(\mathcal{F}_g([\mu_c, \sigma^2_c])) + 1, 
\end{equation}
where $\mathcal{F}_g$ denotes the fully-connnected layers. The Softplus activation ensures non-negativity and the constant shift enforces $\alpha_c > 1$.
Then the calibrated RAW is obtained via:
\begin{equation}
    X_{g} = \alpha_c \cdot X_c.
\end{equation}
For each input, the module learns appropriate per-channel gains to adaptively enhance RAW representations, which are optimized jointly with the downstream model to improve task-specific performance.

\subsubsection{Hierarchical Spatial Attention.} 
Informative content in RAW images is unevenly distributed across spatial locations: some pixels correspond to highly informative structures such as edges or textured regions, while others belong to low-intensity background or flat areas. Moreover, these structures vary in size, with fine details requiring local context and larger regions benefiting from broader contextual information. Treating all pixels equally can weaken important signals and limit the usefulness of the features for downstream tasks. 

To address these challenges, we design a hierarchical spatial attention module that adaptively emphasizes informative pixels across multiple scales, enhancing local features while suppressing irrelevant regions.
Given the feature map processed by the global luminance calibration module $X_g \in \mathbb{R}^{C \times H \times W}$, we first compute channel-aggregated spatial descriptors using both average and max pooling and concatenate them to form a pooled input 
$M = [M_{\text{avg}};M_{\text{max}}] \in \mathbb{R}^{2 \times H \times W}.$
This pooled map is fed into multiple convolutional branches with varying kernel sizes $k \in \{k_1, k_2, \dots\}$, producing single-channel attention maps $A_k$ at different receptive fields. This procedure can be formulated as:
\begin{equation}
\begin{aligned}
    M_{\text{avg}} &= \frac{1}{C} \sum_{c=1}^{C} X_{g,c}, 
    \quad M_{\text{max}} = \max_{c=1,\dots,C} X_{g,c}, \\
    A_k &= \sigma(\text{Conv}_k(M)), \quad k \in \{k_1, k_2, \dots\},
\end{aligned}
\end{equation}
where $\sigma$ denotes the sigmoid activation. To adaptively combine these multi-scale maps, we compute a global descriptor by average pooling the input feature map across spatial dimensions and project it to per-branch weights via a $1*1$ convolution followed by softmax function:
\begin{equation}
\begin{aligned}
    d = \mathcal{G}(X_g) \in \mathbb{R}^{C \times 1 \times 1}, \quad  \\
    w_k = \frac{\exp(\text{Conv}_{1\times1}(d)_k)}{\sum_j \exp(\text{Conv}_{1\times1}(d)_j)},
\end{aligned}
\end{equation}
where $\mathcal{G}$ represents global average pooling. Finally, the fused attention map is obtained as a weighted aggregation of the branch-specific maps and applied to the input feature map via element-wise multiplication:
\begin{equation}
\begin{split}
    A = \sum_k w_k \cdot A_k,& \quad A \in \mathbb{R}^{1 \times H \times W}, \\
   X_s =& X_g \odot A.
\end{split}
\end{equation}
This hierarchical design allows the network to model spatial dependencies across multiple scales while adaptively weighting each scale based on the global context, thereby highlighting task-relevant structures and maintaining computational efficiency.

\subsubsection{Region-Guided Feature Conditioning.} 
To further enable adaptive distribution adjustment, we introduce a region-guided feature conditioning module that allows the network to learn its own spatial partitioning and perform region-specific enhancement. Rather than applying a single transformation across the whole image, the module predicts a set of spatial masks, each corresponding to a learned region, and estimates a modulation weight for every mask. The final output is computed by combining these regionally enhanced features, enabling the network to emphasize content-rich areas while preserving the overall balance of the image. This learned partitioning is entirely data-driven, allowing the model to automatically discover and exploit spatial differences without manual design.

Given the feature processed by the hierarchical spatial attention module $X_s \in \mathbb{R}^{C\times H\times W}$, we first apply a convolutional head $F_m$ to produce mask logits and obtain $K$ groups of spatial masks via Gumbel–Softmax:
\begin{equation}
  M = \mathcal{S_{\tau}}(F_m(X_s), \tau), \quad M \in \mathbb{R}^{B\times K\times H\times W},  
\end{equation}
where $\mathcal{S_{\tau}}$ denotes the Gumbel–Softmax operation with temperature $\tau$, and $\tau$ controls the sharpness of the soft mask selection; in our experiments, we set $\tau=0.1$ to encourage near-discrete masks. Each mask $M_k$ denotes a learned spatial grouping over the image.
To control the degree of enhancement applied to each region, we assign an adaptive scalar weight to every mask.
Concretely, we first obtain a global pooled descriptor from the mask logits and feed it through a small MLP $F_w$ that outputs $K$ mask logits; these are then linearly mapped to a scalar weight $w_k$:
\begin{equation}
  w_k = F_w(\mathcal{P}(F_m(X_s)),  
\end{equation}
where $\mathcal{P}(\cdot)$ denotes global average pooling over spatial dimensions.
Finally, the masks are applied to the input via a weighted power transformation and aggregated:
\begin{equation}
    X_{\text{o}} = \sum_{k=1}^{K} M_k \cdot X_s^{1/w_k}.
\end{equation}
This formulation enables the network to discover spatial partitions and enhance them with learned intensities, yielding fine-grained, context-aware feature conditioning while remaining computationally lightweight.

\begin{figure*}[h]
    \centering
    \includegraphics[width=1.00\linewidth]{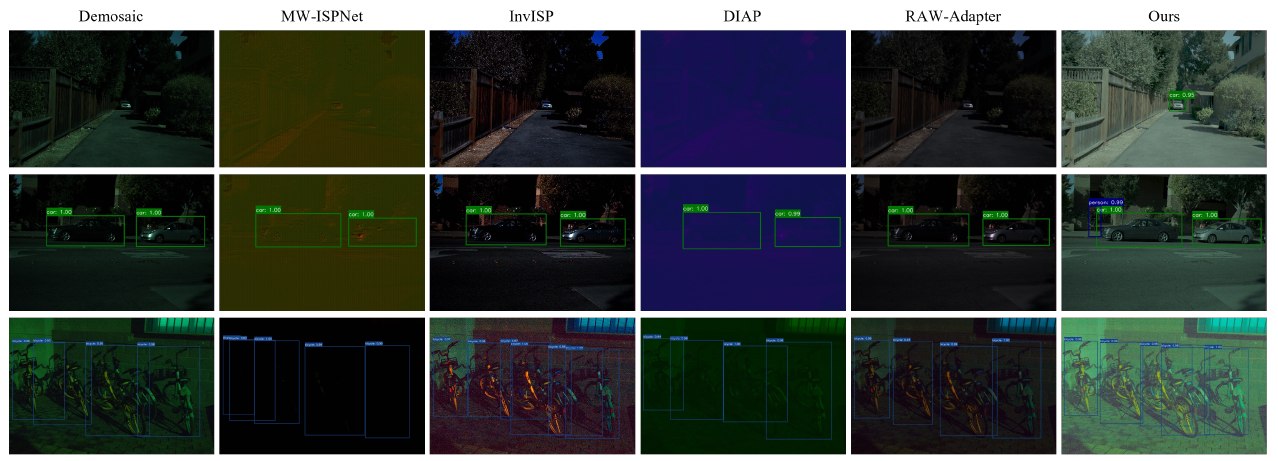}
    \caption{Visual results on the PASCAL RAW dataset \cite{omid2014pascalraw} using RetinaNet \cite{lin2018focallossdenseobject} are shown. The first and second rows present the detection results on the PASCAL RAW dataset, while the third row shows the results on the LOD dataset. Our method demonstrates superior performance in detecting small and occluded objects (Zoom in for best view).}
    \label{vis_det}
\end{figure*}

\begin{table*}[t!]
    \centering
    \setlength{\tabcolsep}{9pt}
    \begin{tabular}{lccc ccc}
        \toprule
        \multirow{2}{*}{Method}
          & \multicolumn{2}{c}{PASCAL RAW}
          & LOD
          & \multirow{2}{*}{Params (M)}
          & \multirow{2}{*}{FLOPs (G)}
          & \multirow{2}{*}{Latency (ms)} \\
        \cmidrule(lr){2-3} \cmidrule(lr){4-4}
          & ResNet-18 & ResNet-50 & ResNet-50 &  &  & \\
        \midrule
        Default ISP          & 88.3 & 89.6 &  58.4 & — & — & — \\
        Demosaic             & 87.7 & 89.2 & 58.5  & — & — & — \\
        MW-ISPNet \cite{ignatov2020aim2020challengelearned} 
                             & 88.9 & 89.6 & 59.4  & 9.14 & 1690.54 & 2425.87 \\
        Karaimer \textit{et al.} \cite{KaraimerB16} 
                             & 88.1 & 89.4 & 54.4  & — & — & — \\
        InvISP  \cite{xing2021invertibleimagesignalprocessing}    
                             & 85.4 & 87.6 & 56.9  & 1.06 & 433.30 & 1584.65 \\
        DIAP \cite{diap}  & 88.5 & 89.7 & 59.5  & 0.08 & 0.23 & 79.70 \\
        RAW-Adapter \cite{cui2024rawadapteradaptingpretrainedvisual}  
                             & 88.7 & 89.7 & 62.1  & 0.76 & 4.02 & 158.01 \\
       \rowcolor{gray!20} Ours  
                             & \textbf{89.9} & \textbf{90.2} & \textbf{63.9}  & \textbf{0.003} & \textbf{0.20} & \textbf{26.43} \\
        \bottomrule
    \end{tabular}
    \caption{Comparison of different methods on the PASCAL RAW dataset \cite{omid2014pascalraw} and the LOD dataset in terms of AP. The best result is highlighted in bold. FLOPs are measured on 640 $\times$ 640 RAW inputs, while inference latency is evaluated on 3840 $\times$ 2160 RAW inputs.}
    \label{tab:comparison_pascal_lod}
\end{table*}

\begin{table}[t]
    \centering
    \setlength{\tabcolsep}{8.0pt}
    \begin{tabular}{l cc cc}
        \toprule
        \multirow{2}{*}{Methods} & \multicolumn{2}{c}{Day} & \multicolumn{2}{c}{Night} \\
        \cmidrule(lr){2-3} \cmidrule(lr){4-5}
         & AP & AP50 & AP & AP50 \\
        \midrule
        Demosaic   & 36.1 & 49.4 & 54.5 & 80.6 \\
        MW-ISPNet \cite{ignatov2020aim2020challengelearned} & 33.6 & 46.5 & 15.9 & 29.6 \\
        InvISP \cite{xing2021invertibleimagesignalprocessing} & 32.8 & 44.7 & 11.9 & 22.5 \\
        DIAP \cite{diap} & 36.2 & 49.9 & 58.5 & 84.3 \\
        RAW-Adapter \cite{cui2024rawadapteradaptingpretrainedvisual} & 35.9 & 49.0 & 45.9 & 69.9 \\
        \rowcolor{gray!20}Ours   & \textbf{38.0} & \textbf{51.6} & \textbf{59.7} & \textbf{84.8} \\
        \bottomrule
    \end{tabular}
    \caption{Comparison of different methods on the ROD dataset \cite{diap} with respect to AP and AP50 for daytime and nighttime settings. \textbf{Bold} indicates the best result.}
    \label{ROD_full}
    \vspace{-1em}
\end{table}

\begin{figure*}[h]
    \centering
    \includegraphics[width=1.00\linewidth]{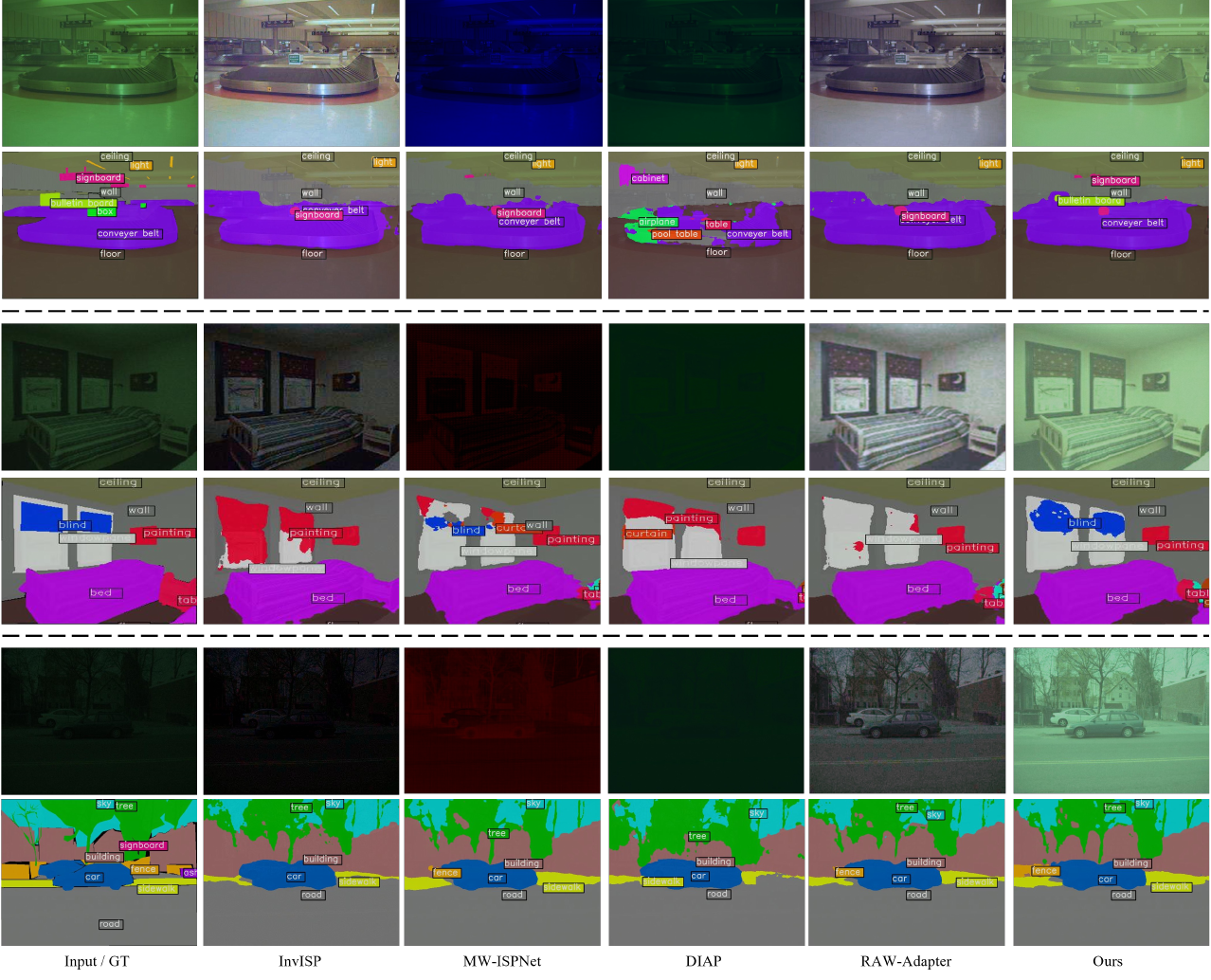}
    \caption{Semantic segmentation results on normal and low-light RAW data on ADE20K dataset \cite{ADE20K}. Our method achieves more accurate
delineation of regions and clearer segmentation results (Zoom in for best view).}
    \label{vis_seg}
    \vspace{-1em}
\end{figure*}

\section{Experiments}

\subsection{Experimental Settings}
\paragraph{Dataset.} We conducted experiments on three real-world RAW detection datasets and one synthetic RAW segmentation dataset, evaluating performance on both daytime and nighttime scenes.

For the object detection task, we use three public RAW object detection datasets: PASCAL RAW \cite{omid2014pascalraw}, LOD \cite{Hong2021Crafting} and ROD \cite{diap}. 
The PASCAL RAW dataset consists of 12-bit RAW images captured exclusively in daytime scenes using a Nikon D3200 DSLR. It contains 4,259 annotated RAW images with three annotated object classes.
The LOD dataset is designed for low-light object detection and was collected using a Canon EOS 5D Mark IV camera. It comprises 2,230 images with a total of 9,726 annotated instances spanning 8 common object classes.
The ROD dataset comprises 24-bit RAW images captured in both daytime and nighttime using a Sony IMX490 sensor. These images retain the high dynamic range of the scene, and therefore pose a substantial challenge for the ISP network. Note that the publicly available ROD dataset includes 4,053 daytime scenes and 12,036 nighttime scenes, covering 5 classes, rather than the 25,207 images and 6 classes as reported in DIAP \cite{diap}.

For the task of object segmentation, we follow RAW-Adapter \cite{cui2024rawadapteradaptingpretrainedvisual}
 to use the ADE20K dataset \cite{ADE20K} to synthesize RAW object segmentation data. ADE20K comprises 27,572 images with annotations for 150 classes. Following the approach of Cui \textit{et al.} \cite{cui2024rawadapteradaptingpretrainedvisual}, we used InvISP \cite{xing2021invertibleimagesignalprocessing} to generate the corresponding RAW dataset, which includes scenes captured under various conditions including daytime and low-light.

\paragraph{Comparison Methods.} To demonstrate the superiority of our approach, we compared our method with the state-of-the-art algorithms. Specifically, default ISP and Karaimer \textit{et al} are the separately designed ISP pipelines, while MW-ISPNet \cite{ignatov2020aim2020challengelearned} and InvISPNet \cite{xing2021invertibleimagesignalprocessing} serve as leading network-based solutions. In addition, DIAP \cite{diap} and RAW-Adapter \cite{cui2024rawadapteradaptingpretrainedvisual} are included as top-performing joint training approaches. We also compare our framework with a downstream model that directly processes the demosaicked images. We build our detection and segmentation  framework based on the open-source codebases DIAP \cite{diap} and RAW-Adapter\cite{cui2024rawadapteradaptingpretrainedvisual}. The following sections provide a detailed explanation of the experimental setup.

\subsection{Experiments on Object Detection}

For the object detection task, we follow the experimental settings of RAW-Adapter \cite{cui2024rawadapteradaptingpretrainedvisual} and DIAP \cite{diap} and conducted experiments using two mainstream object detectors: RetinaNet \cite{lin2018focallossdenseobject} and YOLOX-Tiny \cite{yolox2021}.
All competing approaches, including ours, are initialized with the same pretrained downstream model to ensure a fair comparison.

For the experiments on the PASCAL RAW dataset \cite{omid2014pascalraw}, we adopt RetinaNet with different ResNet backbones \cite{he2015deepresiduallearningimage} (ResNet-18 and ResNet-50). All models are trained on 2 NVIDIA GeForce RTX 4090 GPUs with a batch size of 4 using the SGD optimizer. Training images are randomly cropped to a resolution within the range of (400, 667), and training is conducted for 50 epochs. 
The results are summarized in Table \ref{tab:comparison_pascal_lod}, where we compare our method against demosaicked RAW data, separately designed ISP pipelines \cite{KaraimerB16}, network-based methods \cite{ignatov2020aim2020challengelearned, xing2021invertibleimagesignalprocessing}, and state-of-the-art joint training methods \cite{cui2024rawadapteradaptingpretrainedvisual, diap}. We observe that while separately designed ISP pipelines can improve detection performance, some network-based and joint training methods such as MW-ISPNet \cite{ignatov2020aim2020challengelearned} and RAW-Adapter \cite{cui2024rawadapteradaptingpretrainedvisual} achieve superior results.
Among all compared approaches, our method consistently achieves the best performance across different backbone sizes. Remarkably, our ResNet-18 based model even surpasses previous methods utilizing ResNet-50 backbones. Furthermore, as shown in Table \ref{tab:comparison_pascal_lod}, our approach is highly efficient, requiring only 3K parameters, and can process a $3840 \times 2160$ image under 27 ms, while delivering state-of-the-art accuracy.

Experimental results on the LOD dataset \cite{Hong2021Crafting} are shown in Table \ref{tab:comparison_pascal_lod}, we train our model using RetinaNet with a ResNet-50 backbone. The batch size is set to 4, and training is conducted for 35 epochs. As shown in the results, our method significantly outperforms all other approaches, achieving the best performance under low-light conditions.

The visualizations of the PASCAL RAW dataset \cite{omid2014pascalraw} and the LOD dataset \cite{Hong2021Crafting} are presented in Figure \ref{vis_det}. We present detection results under both daytime and nighttime conditions. As shown, our method is able to accurately identify a greater number of small and occluded objects than competing approaches.

To further assess the robustness of our model, we conducted experiments on the ROD dataset \cite{diap}, which contains both daytime and nighttime scenes with high dynamic range. The results are presented in Table \ref{ROD_full}. We observe that many existing methods struggle to maintain strong performance under such challenging conditions, whereas our approach consistently achieves the best results across all comparisons.

\begin{table}[htbp!]
    \centering
    \setlength{\tabcolsep}{15pt}
    \begin{tabular}{cccc}
        \toprule 
        \multirow{2}{*}{Methods} & \multicolumn{2}{c}{mIOU} \\
        \cmidrule{2-3} & normal &  dark\\
        \midrule
        Demosaic & 34.27 & 25.49 \\
        MW-ISPNet \cite{ignatov2020aim2020challengelearned} & 35.27& 25.06\\
        InvISP \cite{xing2021invertibleimagesignalprocessing} & 34.02& 21.07\\
        DIAP \cite{diap} & 30.54 & 23.74\\
        RAW-Adapter \cite{cui2024rawadapteradaptingpretrainedvisual} & 34.72 & 25.06\\
        \rowcolor{gray!20} Ours & \textbf{36.29} & \textbf{26.77}  \\
        \bottomrule
    \end{tabular}
    
    \caption{%Comparison of different methods in terms of AP across different training data proportions on the PASCAL RAW dataset
    Comparison of different methods on the synthetic ADE20K dataset \cite{ADE20K} with respect to mIOU. \textbf{Bold} indicates the best result.
    \label{seg}
    }
    \vspace{-1.5em}
\end{table}

\subsection{Experiments on Semantic Segmentation}

For the semantic segmentation task on the ADE20K dataset \cite{ADE20K}, we use Segformer \cite{xie2021segformer} as the decoder and MITB0 \cite{xie2021segformer} as the backbone. To ensure a fair comparison, all models are trained with the same settings as those used in RAW-Adapter \cite{cui2024rawadapteradaptingpretrainedvisual}. Training images are cropped to 512 × 512, and the number of training iterations is set to 80,000.

The comparison results are summarized in Table \ref{seg}. While certain joint training methods achieve competitive performance under normal lighting conditions, their effectiveness drops considerably in low-light scenarios. MW-ISPNet \cite{ignatov2020aim2020challengelearned} and RAW-Adapter \cite{cui2024rawadapteradaptingpretrainedvisual} deliver stronger results, but at the expense of either substantial computational overhead or increased bandwidth requirements. In contrast, our framework consistently exhibits superior adaptability across diverse environments and achieves the best overall performance in all cases by a significant margin.

Visualization results of segmentation are shown in Figure \ref{vis_seg}. The first two rows display segmentation results under normal lighting conditions, while the latter two rows present results in low-light scenarios. Our method consistently outperforms other approaches, delivering more accurate region delineation and clearer segmentation outputs.

\begin{table}[htbp]
  \centering
  \setlength{\tabcolsep}{7pt}
  \vspace{0.25em}
  \begin{tabular}{@{}ccccc@{}}
    \toprule
    \multirow{2}{*}{GLC} & 
    \multirow{2}{*}{HSA} & 
    \multirow{2}{*}{RGFC} & 
    \multicolumn{2}{c}{AP} \\
    \cmidrule(lr){4-5}
    & & & ROD & LOD \\
    \midrule
    -- & -- & -- & 36.1 & 58.5 \\
    \checkmark & -- & -- & 36.3 (+0.2) & 60.4 (+1.9) \\
    \checkmark & \checkmark & -- & 36.8 (+0.7) & 63.0 (+4.5) \\
    \checkmark & \checkmark & \checkmark & 38.0 (+1.9) & 63.9 (+5.4) \\
    \bottomrule
  \end{tabular}
  \caption{Ablation study of different modules in TA-ISP evaluated on the ROD \cite{diap} and LOD \cite{Hong2021Crafting} datasets.}
  \label{ablation}
  \vspace{-1em}
\end{table}

\subsection{Ablation Study}
\paragraph{Modules of TA-ISP.}
In this section, we present ablation studies to assess the effectiveness of the proposed TA-ISP modules: Global Luminance Calibration (GLC), Hierarchical Spatial Attention (HSA), and Region-Guided Feature Conditioning (RGFC). Experiments are conducted on the ROD dataset \cite{diap} for daytime scenes and the LOD dataset \cite{Hong2021Crafting} for nighttime scenes, using YOLOX-Tiny \cite{yolox2021} and RetinaNet with a ResNet-50 backbone, respectively. As summarized in Table \ref{ablation}, the results demonstrate that our modules consistently improve performance across daytime, nighttime, and high dynamic range scenarios.

\begin{table}[htbp!]
    \centering
    \setlength{\tabcolsep}{11pt}
    \begin{tabular}{cccc}
        \toprule 
        \multirow{2}{*}{Methods} & \multicolumn{3}{c}{Percentage} \\
        \cmidrule{2-4} & 10$\%$ & 25$\%$ & 50$\%$\\
        \midrule
        Demosaic & 73.9 & 82.0& 85.2 \\
        MW-ISPNet \cite{ignatov2020aim2020challengelearned} & 68.4 & 81.3& 84.9\\
        InvISP \cite{xing2021invertibleimagesignalprocessing} & 66.8 & 79.5 &  80.1\\
        DIAP \cite{diap} & 71.5 & 83.1& 86.3\\
        RAW-Adapter \cite{cui2024rawadapteradaptingpretrainedvisual}  & 70.1 & 81.5& 84.5\\
        \rowcolor{gray!20} Ours  & \textbf{74.1}  & \textbf{86.4}& \textbf{88.9}\\
        \bottomrule
    \end{tabular}
    
    \caption{ Comparison of ISP methods on PASCAL RAW \cite{omid2014pascalraw} under limited training data.
    \label{limited}
    }
    \vspace{-1em}
\end{table}

\paragraph{Limited Datasets.}
To further evaluate the effectiveness of our TA-ISP under limited training data, we conduct experiments on the PASCAL RAW dataset \cite{omid2014pascalraw} using RetinaNet with a ResNet-18 backbone. Specifically, we randomly sample 10$\%$, 25$\%$, and 50$\%$ of the training set (2129 images in total). The results are summarized in Table~\ref{limited}. We observe that when training data is scarce, most ISP methods tend to degrade significantly and even perform worse than the demosaicing baseline. In contrast, our TA-ISP consistently outperforms demosaicing across all settings. More importantly, with only 25$\%$ or 50$\%$ of the training data, TA-ISP already surpasses all competing methods that are trained with 50$\%$ or even 100$\%$ of the data, respectively. This demonstrates not only the robustness of our approach in data-limited conditions but also its superior data efficiency, enabling strong downstream performance with substantially fewer labeled samples.

\begin{table}[htbp!]
    \centering
    \setlength{\tabcolsep}{15pt}
    \begin{tabular}{ccc}
        \toprule 
        Methods & AP & AP50 \\
        \midrule
        Demosaic & 41.6& 56.6 \\
        MW-ISPNet \cite{ignatov2020aim2020challengelearned} & 38.3& 51.5\\
        DIAP \cite{diap}  & 41.7& 55.7\\
        RAW-Adapter \cite{cui2024rawadapteradaptingpretrainedvisual} & 40.6 & 53.7 \\
        \rowcolor{gray!20} Ours&  \textbf{42.9}& \textbf{57.4}\\
        \bottomrule
    \end{tabular}
    
    \caption{Evaluation of different ISP methods on the ROD dataset \cite{diap} with YOLOX-L \cite{yolox2021}.
    \label{size}
    }
    \vspace{-1.5em}
\end{table}

\paragraph{Model Size.}
To examine the influence of model size on RAW sensor data, we evaluate YOLOX with varying parameter configurations on the ROD dataset \cite{diap}. Specifically, we adopt YOLOX-L for joint training with mainstream ISP methods. As reported in Table \ref{size}, our approach consistently surpasses competing methods as the model size increases.

\section{Conclusion}

In this paper, we introduced TA-ISP, a task-aware, lightweight RAW to RGB framework that produces task-oriented representations for pretrained vision models. TA-ISP factorizes image adaptation into compact, multi-granularity pixel-modulation modules operating at global, regional, and pixel scales, enabling image-wide luminance/color calibration, spatially coherent regional adjustments, and fine-grained per-pixel corrections. By predicting small, efficient modulation operators instead of relying on dense convolutions, our design expands the class of spatially-varying transforms that can be represented while keeping parameter count, memory footprint, and inference latency tightly constrained—making TA-ISP practical for resource-limited deployments. Extensive experiments on multiple RAW-domain detection and segmentation benchmarks under both daytime and nighttime conditions demonstrate that TA-ISP consistently and substantially improves downstream task performance compared with prior ISP methods, all with only minimal computational overhead. We believe TA-ISP represents a practical middle ground between heavyweight learned ISPs and simple parameter-tuning schemes, and we hope it will inspire further work on compact, task-driven imaging front-ends.
{
    \small
    \bibliographystyle{ieeenat_fullname}
    \bibliography{main}
}

\clearpage

\setcounter{page}{1}
\setcounter{section}{0}
\renewcommand{\thesection}{\Alph{section}}
\maketitlesupplementary
% \onecolumn

\section{Implementation Details}
In this section, we provide additional implementation details for our experiments.

\paragraph{Object Detection}
For object detection experiments on the ROD dataset \cite{diap}, all ISP models were trained jointly with a pretrained YOLOX-Tiny detector \cite{yolox2021}. We randomly selected 90$\%$ of the images for training and used the remaining 10$\%$ for testing. The batch size was set to 16, and the training images were resized to $640 \times 640$. The joint models were trained for 300 epochs using the SGD optimizer with an initial learning rate of $2.5 \times 10^{-3}$ on two NVIDIA GeForce RTX 4090 GPUs.

For experiments on the PASCAL RAW dataset \cite{omid2014pascalraw}, we adopted RetinaNet \cite{lin2018focallossdenseobject} with two different backbones, ResNet-18 \cite{he2015deepresiduallearningimage} and ResNet-50. Training was conducted for 50 epochs, with images randomly cropped to a resolution of $(400, 667)$. The dataset was split into training and testing sets according to the official PASCAL RAW protocol \cite{omid2014pascalraw}. All models were trained on a single NVIDIA GeForce RTX 4090 GPU using the SGD optimizer. The initial learning rate was set to $5 \times 10^{-3}$ for the ResNet-18 backbone and $2 \times 10^{-3}$ for the ResNet-50 backbone.

For experiments on the LOD dataset \cite{Hong2021Crafting}, we employed RetinaNet with a ResNet-50 backbone. The batch size was set to 4, and the models were trained for 35 epochs using the SGD optimizer with an initial learning rate of $5 \times 10^{-3}$. All training was performed on a single NVIDIA GeForce RTX 4090 GPU.

\paragraph{Segmentation}
For semantic segmentation experiments on the synthetic ADE20K dataset \cite{ADE20K}, we followed the settings of Cui \textit{et al.} \cite{cui2024rawadapteradaptingpretrainedvisual}. Training images were cropped to $512 \times 512$, and the number of training iterations was set to 80,000. The models were trained on four NVIDIA GeForce RTX 4090 GPUs with Adam optimizer.

\section{Traditional Image Signal Processor}
In a conventional image signal processing (ISP) pipeline, raw sensor measurements are transformed into display-ready RGB images through a sequence of modular operations. Below we summarize the typical modules and their primary functions, presented in an approximate processing order.

\begin{enumerate}
  \item \textbf{Raw preprocessing.} Raw sensor data are first corrected for sensor bias by subtracting the black level and optionally applying gain/offset corrections. Linearization compensates for non-linear ADC responses so subsequent operations operate in a scene-linear radiometric domain.
  \item \textbf{Defective-pixel correction and hot-pixel removal.} Pixels known to be dead or noisy are detected and replaced, typically by interpolation from neighboring pixels or median filtering, to prevent localized artifacts from propagating through the pipeline.
  \item \textbf{Demosaicing.} Demosaicing reconstructs full RGB color at each pixel from the mosaiced sensor pattern (e.g., Bayer). Algorithms estimate missing color components while aiming to preserve edges and minimize color zippering or bleeding.
  \item \textbf{Denoising.} Denoising reduces photon and readout noise present in the linear raw domain. Effective denoising preserves fine texture and edges while suppressing stochastic noise, often using spatial, temporal (for burst), or frequency-domain filtering.
  \item \textbf{White balance.} White balance scales the three color channels to compensate for the scene’s illumination color so that neutral surfaces appear neutral. This operation is typically performed in the linear domain and may be driven by metadata, statistics, or automatic estimation.
  \item \textbf{Color correction.} A color correction matrix maps camera-native RGB to the target colorimetry (e.g., linear scene-referred RGB or display primaries). The 3$\times$3 matrix corrects systematic spectral differences between sensor responses and the desired color space.
  \item \textbf{Tone mapping and dynamic range compression.} Tone mapping converts scene-referred linear intensities to display-referred values, compressing high dynamic range to the limited display range while preserving perceptual contrast. This step may use parametric curves or more complex mapping functions.
  \item \textbf{Gamma correction.} Nonlinear gamma encoding (e.g., the sRGB transfer) is applied to map linear scene luminance into a perceptually uniform display-referred space, improving visual fidelity on standard displays.
  \item \textbf{Sharpening and detail enhancement.} Sharpening boosts high-frequency components to improve perceived crispness. Methods typically apply edge-aware unsharp masking or high-frequency residuals while mitigating halo artifacts.
  \item \textbf{Compression and encoding.} Finally, the rendered RGB images are quantized and encoded (e.g., JPEG, HEIF) for storage or transmission, including any chroma subsampling and metadata embedding.
\end{enumerate}

\begin{table}[h]
	\centering
        %\caption{Comparison of AdaptiveISP on LOD dataset.}
        \setlength{\tabcolsep}{14pt}  
        \begin{tabular}{lcc}
            \toprule
            Number & PASCAL RAW & LOD  \\
            \midrule
            n=4  & 88.6 &  62.2\\
            n=8  & 89.0 &  63.3\\
            n=16 & 89.9 &  63.9\\
            n=32 & 88.4 &  62.0\\
            \bottomrule
        \end{tabular}
        \caption{Ablation on the number of mask layers on PASCAL RAW dataset \cite{omid2014pascalraw} and LOD dataset \cite{Hong2021Crafting}.}
        \label{num}
\end{table}

\section{Ablation on the Number of Mask Layers}
We conducted experiments on the PASCAL RAW dataset \cite{omid2014pascalraw} using RetinaNet with ResNet-18 and on the LOD dataset \cite{Hong2021Crafting} using RetinaNet with ResNet-50 to determine the optimal number of mask layers. Specifically, we evaluated models with 4, 8, 16, and 32 layers, as summarized in Table \ref{num}. The results show that increasing the number of mask layers generally improves performance on both daytime and nighttime datasets. However, when the number of layers becomes too large, performance drops significantly. Based on these observations, we set the number of mask layers to 16 in this work.

\section{More Visualization Results}
\begin{figure*}[t]
    \centering
    \includegraphics[width=1.00\linewidth]{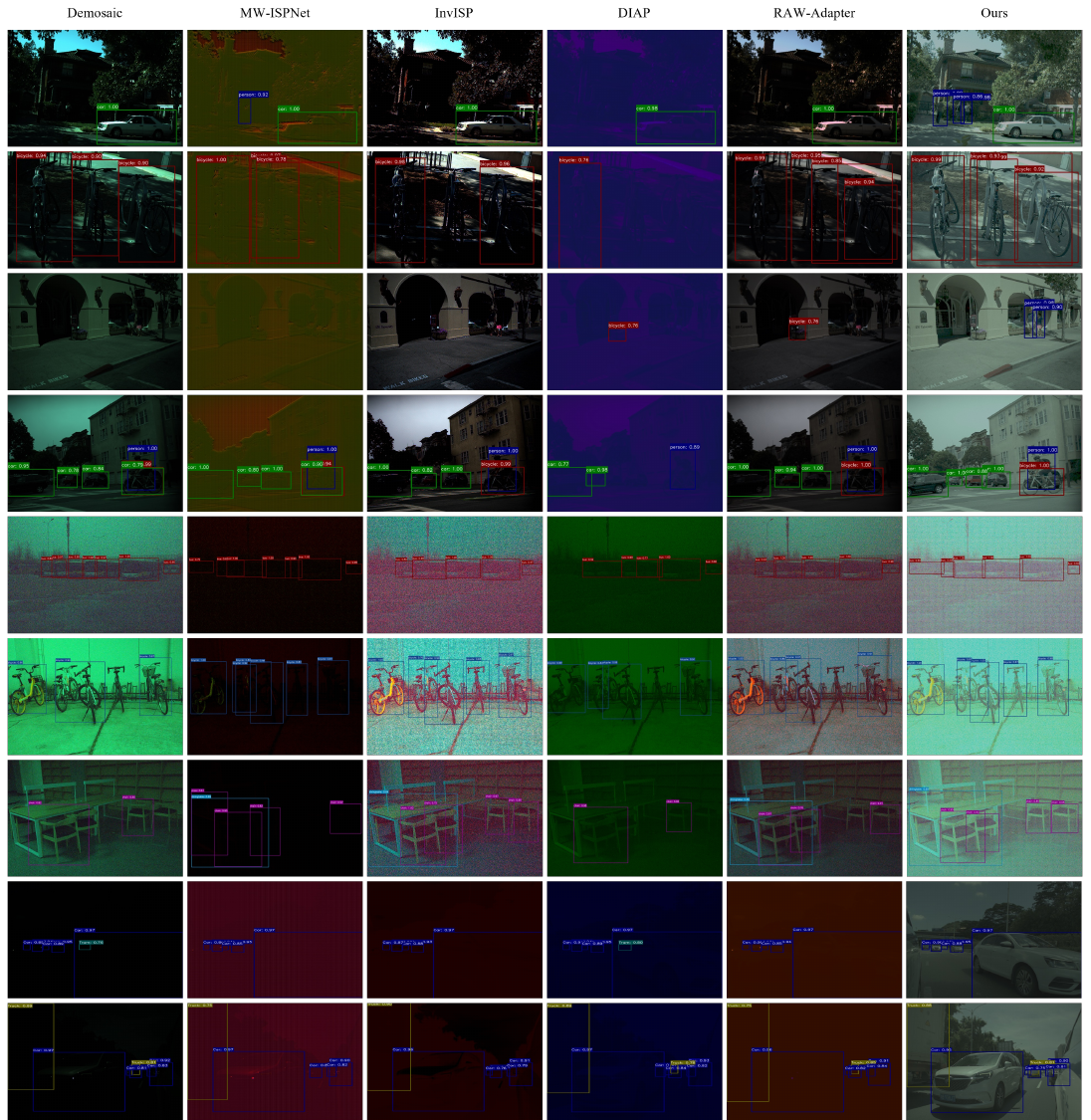}
    \caption{Visual results on PASCAL RAW dataset \cite{omid2014pascalraw} , LOD dataset \cite{Hong2021Crafting} and ROD dataset \cite{diap}. (Zoom in for best view).}
    \label{sup_vis_det}
\end{figure*}

\begin{figure*}[t]
    \centering
    \includegraphics[width=1.00\linewidth]{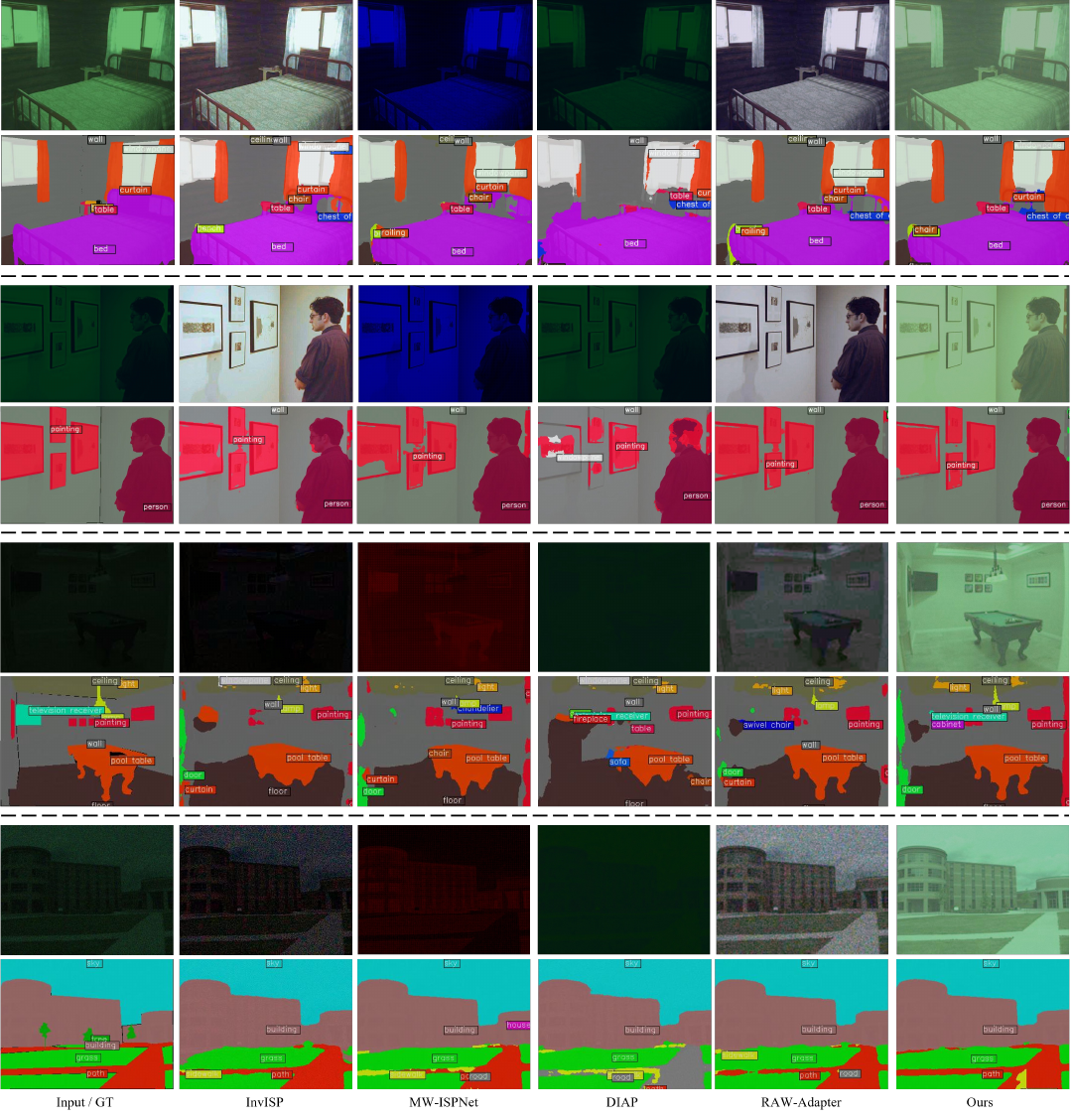}
    \caption{Semantic segmentation results on normal and low-light RAW data on ADE20K dataset \cite{ADE20K}. (Zoom in for best view).}
    \label{sup_vis_seg}
\end{figure*}

In this section, we present additional visualization results. We compare our method with Demosaic, InvISP \cite{xing2021invertibleimagesignalprocessing}, MW-ISPNet \cite{ignatov2020aim2020challengelearned}, DIAP \cite{diap}, and RAW-Adapter \cite{cui2024rawadapteradaptingpretrainedvisual}. The detection results are shown in Figure \ref{sup_vis_det}, where rows 1–4 correspond to the PASCAL RAW dataset \cite{omid2014pascalraw}, rows 5–7 correspond to the LOD dataset \cite{Hong2021Crafting}, and rows 8–9 correspond to the ROD dataset \cite{diap}. The segmentation results are illustrated in Figure \ref{sup_vis_seg}, where we provide a side-by-side comparison with the aforementioned methods.

\clearpage

\end{document}